\documentclass[letterpaper, 10 pt, conference]{ieeeconf_icra}  % Comment this line out if you need a4paper

\IEEEoverridecommandlockouts                              % This command is only needed if 
\usepackage{graphics} % for pdf, bitmapped graphics files
\usepackage{epsfig} % for postscript graphics files
\usepackage{mathptmx} % assumes new font selection scheme installed
\usepackage{times} % assumes new font selection scheme installed
\usepackage{amsmath} % assumes amsmath package installed
\usepackage{amssymb}  % assumes amsmath package installed
\usepackage{subcaption}
\usepackage[acronym]{glossaries}
\usepackage{gensymb}
\usepackage{cite}
\usepackage{multirow}
\usepackage{booktabs}
\usepackage{pdfpages}
\usepackage{graphicx}
\usepackage{extdash}

\usepackage{pifont}% http://ctan.org/pkg/pifont

\makeatletter
\let\NAT@parse\undefined
\makeatother

\usepackage{hyperref}

\newacronym{ACW}{ACW}{Anti-clockwise}
\newacronym{CW}{CW}{Clockwise}
\newacronym{KDE}{KDE}{Kernel Density Estimation}
\newacronym{DNN}{DNN}{Deep Neural Network}
\newacronym{RNN}{RNN}{Recurrent Neural Network}
\newacronym{CNN}{CNN}{Convolutional Neural Network}
\newacronym{CfC}{CfC}{Closed-form Continuous-time Neural Network}
\newacronym{LTC}{LTC}{Liquid Time-Constant}
\newacronym{LRC}{LRC}{Liquid Resistance Liquid Capacitance Network}
\newacronym{LSTM}{LSTM}{Long Short-Term Memory}
\newacronym{NCP}{NCP}{Neural Circuit Policy}
\newacronym{MSE}{MSE}{Mean Squared Error}
\newacronym{VESC}{VESC}{Vedder Electronic Speed Controller}
\newacronym{IMU}{IMU}{Inertial Measurement Unit}
\newacronym{MLP}{MLP}{Multi-layer Perceptron}
\newacronym{MP}{MP}{McCulloch–Pitts}
\newacronym{E2E}{E2E}{End-to-end}
\newacronym{DRL}{DRL}{Deep Reinforcement Learning}
\newacronym{FC}{FC}{Fully-Connected}

\title{\LARGE \bf
Depth Matters: Multimodal RGB-D Perception\\for Robust Autonomous Agents
}

\author{Mihaela-Larisa Clement$^{*1}$, M\'{o}nika Farsang$^{*1}$, Felix Resch$^{1}$, Mihai-Teodor Stanusoiu$^{1}$, Radu Grosu$^{1}$% <-this % stops a space
\thanks{* denotes equal contribution}
\thanks{$^{1}$CPS, Technische Universit\"{a}t Wien (TU Wien), Austria}%
\thanks{E-mail of corresponding author: mihaela-larisa.clement@tuwien.ac.at}%
}

\begin{document}

\maketitle
\thispagestyle{empty}
\pagestyle{empty}

%%%%%%%%%%%%%%%%%%%%%%%%%%%%%%%%%%%%%%%%%%%%%%%%%%%%%%%%%%%%%%%%%%%%%%%%%%%%%%%%
\begin{abstract}
% draft version%
Autonomous agents that rely purely on perception to make real-time control decisions require efficient and robust architectures. In this work, we demonstrate that augmenting RGB input with depth information significantly enhances our agents' ability to predict steering commands compared to using RGB alone. We benchmark lightweight recurrent controllers that leverage the fused RGB-D features for sequential decision-making. To train our models, we collect high-quality data using a small-scale autonomous car controlled by an expert driver via a physical steering wheel, capturing varying levels of steering difficulty. Our models were successfully deployed on real hardware and inherently avoided dynamic and static obstacles, under out-of-distribution conditions.
%showcasing their robustness across multiple driving conditions. 
Specifically, our findings reveal that the early fusion of depth data results in a highly robust controller, which remains effective even with frame drops and increased noise levels, without compromising the network’s focus on the task.
% Furthermore, our findings reveal that our approach not only produces smoother and more reliable steering predictions but also leads to faster and more consistent reaction times than a human driver.
% to maintain efficiency suitable for real-world deployment 
%Our experiments show that models trained with depth-aware multimodal input achieve xxx lower steering error compared to RGB-only baselines. 

\end{abstract}
%%%%%%%%%%%%%%%%%%%%%%%%%%%%%%%%%%%%%%%%%%%%%%%%%%%%%%%%%%%%%%%%%%%%%%%%%%%%%%%%
\section{INTRODUCTION}
%This is 'my favorite' way of making introductions that I learned in one fo the PhD seminars I attended
% One sentence or short paragraph about what the paper is
% about and a mini-summary of your main results. (1 par)
% • Why is this problem important, who has attempted to solve
% it, which solutions exist so far, why aren‘t they sufficient,
% why are we left with a dilemma, what are the main
% obstacles, why would it be useful to fight them. What would
% be the benefit of a (better) solution. (1/2 page)
% • Elaboration on single obstacles and how you overcome
% them.
%
% • Short description of results and their advantage and
% usefulness in chronological order.
%
% • Some remarks about the nontriviality of the solution
% • Introducing the important problem you solve, give
% some (possibly simplified) definitions (1 par)
% • Further related work and a sentence about future
% plans
% • Summary of main results in form of a dotlist
Contemporary approaches to autonomous driving agents use both unimodal and multimodal input for machine learning models in order to provide accurate controls. However, when deploying the agents in the real world, the sim-to-real gap becomes an issue. This is caused by noise affecting sensor readings and computational delays from heavy processing that can significantly impair agent performance.

On the one hand, unimodal RGB input has been successfully used in real-time applications in prior work~\cite{Lechner_Hasani_Amini_Henzinger_Rus_Grosu_2020}, where a hybrid architecture of convolutional-head with biologically inspired \gls{LTC}~\cite{hasani2021liquid} networks was used. They demonstrated exceptional performance in autonomous lane-keeping tasks by directly mapping raw RGB camera inputs to steering commands, outperforming conventional deep networks in both interpretability and resilience to sensor noise. \textit{One of the goals is to replicate the model architecture used here} as closely as possible, with the necessary adjustments. However, it is known that differential equation (DE) solvers used in their model can cause longer computation times. In recent work,~\cite{Hasani_2022} introduced \gls{CfC}, a variant of the ODE-based continuous LTC network. Another efficient extension is by introducing a liquid capacitance term into the model~\cite{farsang2024liquid} on top of saturation~\cite{farsang2024learning}, called \gls{LRC}. These solutions can be beneficial in lightweight hardware setups, ensuring faster reaction times, which we implement and test in this paper.

\begin{figure}[hbt!]
    \centering
    \includegraphics[width=0.99\linewidth]{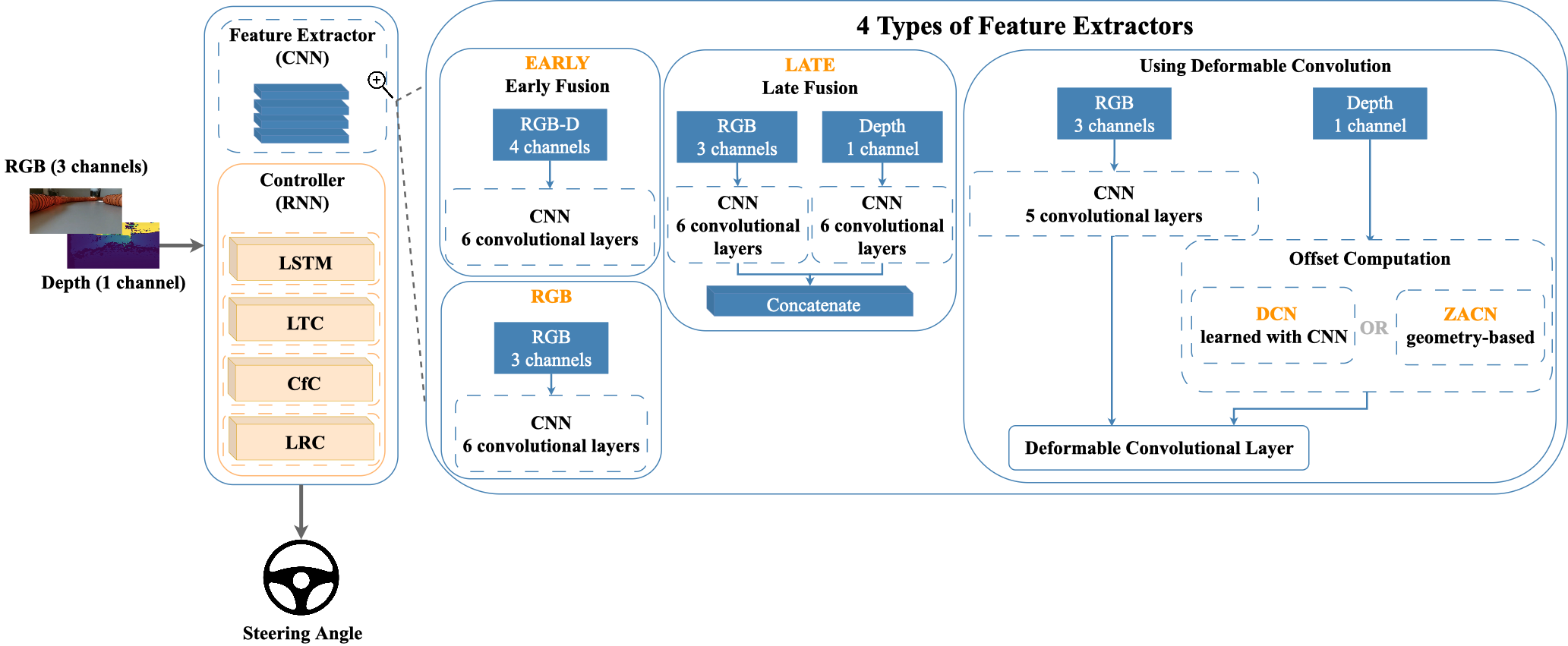}
    \caption{High-level overview of the model architectures.}
    \label{fig:stackdiagram}
    \vspace{-4ex}
\end{figure}

On the other hand, unimodal approaches face inherent limitations in complex scenarios where depth perception is critical for obstacle avoidance and path planning. To tackle this challenge, we employ multimodal sensory fusion. RGB-D input is used in simulation both in~\cite{Mújica-Vargas2024} and ~\cite{MultimodalEnd-to-EndAutonomousDriving}. Although such approaches demonstrate success in simulated environments, their applicability to real-world scenarios remains unexplored, especially on small-scale platforms like \textit{roboracer}~\cite{f1tenth}. This paper seeks to \textit{bridge that gap, both as a real-world application, and with more lightweight architectures}.

In addition, the fusion of RGB-D information has been widely researched in computer vision, mainly using Convolutional Neural Networks (CNN)\cite{lecuncnn}. A widely used approach is introducing depth information as a new channel to the input of 2D convolutions, which yields the best results in~\cite{MultimodalEnd-to-EndAutonomousDriving}. Alternatively, the authors of~\cite{depthawarecnn} use depth information in the computation of Deformable CNNs~\cite{deformcnn, zhu2018deformableconvnetsv2deformable}, which aim to enhance spatial flexibility. Their approach leads to more accurate object classification than with RGB alone. We aim to test these more advanced approaches not in computer vision but in a navigation task.

This paper aims to investigate the use of \mbox{RGB-D} data for navigating in complex scenarios. In this way, we can achieve optimized multimodal models for the small-scale autonomous system \textit{roboracer} with fast reaction times and behavior comparable to that of an experienced human driver.

In summary, our contributions in this paper are as follows:
\begin{itemize}
    \item We show depth information is crucial, RGB alone is not enough for robust navigation, as RGB-only agents fail in scenarios with obstacles, decision points, or sharp turns while RGB-D agents succeed.
    \item A systematic comparison of RGB-D fusion techniques, evaluating depth-as-channel (early, late fusion) and depth-aware deformable CNNs (DCN~\cite{deformcnn} and ZACN\cite{wu2020depthadaptedcnnrgbdcameras}) to optimize feature extraction for perception.
    \item Lightweight recurrent controller benchmarking for efficiency, using features extracted from our multimodal vision block from the previous point for control.
    \item The first deployment of a depth-enhanced network on a \textit{roboracer} vehicle, achieving a smooth and noise-robust control and zero-shot adaptation to unseen scenarios such as static and dynamic obstacles and intersections.
\end{itemize}

% • Roadmap „This paper is organized as follows...“
% The rest of the paper is organized as follows. Section~\ref{sec:related_work} reviews prior work on multimodal perception and autonomous driving architectures. In Section~\ref{sec:methods}, we describe our hardware configuration and dataset collection process. Section~\ref{sec:model_architectures} summarizes our proposed model architectures integrating RGB-D fusion. In Section~\ref{sec:results}, we evaluate the impact of depth-aware feature extraction via vision fusion heads paired with recurrent controllers for sequential steering prediction.
% Finally, we discuss our results in Section~\ref{sec:conclusion}.

\section{RELATED WORK}\label{sec:related_work}
In this section, we review recent studies in autonomous steering control using deep-learning methods, focusing on RGB-based approaches, depth modality fusion, as well as state-of-the-art methods for the \textit{roboracer} platform. We highlight key advancements, limitations in real-world deployment and the gap in efficient resource-constrained solutions.
\subsection{Unimodal RGB Input}
% RGB:
% 1. End to End Learning for Self-Driving Cars
% 2. Neural circuit policies enabling auditable autonomy
% 3. A Virtual End-To-End Learning System for Robot Navigation Based on Temporal Dependencies
% 4. Optical Flow Matters: an Empirical Comparative Study on Fusing Monocular Extracted Modalities for Better Steering
A critical achievement in the context of end-to-end autonomous driving is~\cite{DBLP:journals/corr/BojarskiTDFFGJM16}, in which a \gls{CNN} trained on RGB input successfully predicted steering commands on a drive-by-wire vehicle. Although it is shown the model learned useful road features, it remained impossible to assess which segments of the network contributed to its decisions due to the large scale of the network.

Alternatively,~\cite{9144508} tackled the limitations of traditional \gls{CNN}-based steering models by adding temporal awareness via \gls{LSTM}\cite{lstm}. After training on synthetic RGB images, the model extracted features effectively and navigated obstacle-free scenarios, but real-world deployment was not tested.

Addressing the issues of interpretability of traditional deep learning methods, the authors of~\cite{Lechner_Hasani_Amini_Henzinger_Rus_Grosu_2020} introduced a compact bio-inspired neural controller, called \gls{NCP}. Their model significantly outperformed purely \gls{CNN} methods in hardware deployment, using RGB data.

Another study extracts optical flow from RGB frames to improve steering estimation~\cite{makiyeh2024opticalflowmattersempirical}. By employing both early and hybrid fusion techniques with CNN and \gls{RNN}'s, they demonstrated that incorporating optical flow significantly reduces steering estimation error compared to RGB-only methods in dataset-only evaluation. Nevertheless, the results are limited to dataset experiments, without on-board evaluation.

\subsection{Multimodal Input}
% rgbd: early fusion and late fusion examples
% rgbd and steering
% RGB-D:
% 1. RGB-D Convolutional Recurrent Neural Network to Control Simulated Self-driving Car
% 2. Multimodal End-to-End Autonomous Driving
% 	RGB & event camera:
% 	1. multimodal Fusion for Sensorimotor Coordination in Steering Angle Prediction
Although several studies have explored different approaches to improve the performance of autonomous driving models by leveraging RGB-D information, many of them lack real-world testing to confirm practical viability.

Notably,~\cite{Mújica-Vargas2024} presents a large \gls{CNN}-LSTM trained by fusing camera and LiDAR data. Their results accentuate the importance of multimodal perception for enhancing autonomy, outperforming RGB-only models. However, the model was not evaluated in a real-world driving scenario, leaving its practical deployment untested.

Similarly, another study investigates the impact of multimodal perception wit a branched CNN-MLP architecture, focusing on fusion techniques~\cite{MultimodalEnd-to-EndAutonomousDriving}. Using the CARLA simulator, the authors compare the performance of single-modality (RGB/depth) and multimodality. Their results indicate that early fusion multimodality outperforms single-modality models. However, real-world validation was not performed, limiting its practical insights.

%Both works highlight the advantages of incorporating depth information into autonomous driving models. While the first study focuses on improving RCNN architectures and evaluating their impact on vehicle autonomy, the second study provides a comprehensive analysis of different fusion techniques in an end-to-end learning paradigm. These findings reinforce the growing consensus that multimodal sensor fusion, particularly RGB-D integration, enhances perception and decision-making in autonomous driving systems.

% Beyond traditional vision-based methods, the role of event cameras in autonomous navigation has also been explored. A recent study introduced DRFuser, a multimodal fusion network that combines RGB and event-based vision using a self-attention mechanism within an encoder-decoder architecture~\cite{munir2022multimodalfusionsensorimotorcoordination}. By leveraging self-attention layers, the model effectively integrates spatial and temporal dependencies across both modalities, achieving good performance in dataset evaluation.

\subsection{\textit{roboracer}: Learning Approaches and Sensor Modalities} %State-of-the-Art \textit{roboracer} Research: Learning Approaches and Sensor Modalities
% autonomous \textit{roboracer} 
% \textit{roboracer}:
% Deep reinforcement learning (DRL)
% 1. Bypassing the Simulation-to-reality Gap: Online Reinforcement Learning using a Supervisor
% 2. Safe reinforcement learning for high-speed autonomous racing
% 3. Comparing deep reinforcement learning architectures for autonomous racing
% Supervised Learning:
% 1. TinyLidarNet: 2D LiDAR-based End-to-End Deep Learning Model for \textit{roboracer} Autonomous Racing
% 2. Assessing the Robustness of LiDAR, Radar and Depth Cameras Against Ill-Reflecting Surfaces in Autonomous Vehicles: An Experimental Study RGB and steering
\gls{E2E} deep learning approaches have been widely explored using the \textit{roboracer} platform, with multiple works leveraging \gls{DRL} for control~\cite{evans2023bypassingsimulationtorealitygaponline, EVANS2023107, EVANS2023100496}. However, most of these studies rely on the available simulators: the F1Tenth 2D simulator~\cite{pmlr-v123-o-kelly20a} or the ROS/Gazebo-based variant~\cite{9216949}, which do not provide realistic rendering for RGB-D cameras. In particular, the Gazebo-based camera models fail to capture the visual complexity of real-world environments, making it difficult to evaluate complex pipelines in simulation. This motivates the need for real-world data when studying robotic controllers. In contrast, supervised learning-based approaches for \textit{roboracer} are relatively unexplored, with \mbox{TinyLidarNet}~\cite{zarrar2024tinylidarnet2dlidarbasedendtoend} being a recent contribution. \mbox{TinyLidarNet} introduces a lightweight LiDAR-based \gls{CNN} model, which maintains fast inference efficiency on low-end microcontrollers. However, its reliance on LiDAR raises concerns about robustness in challenging environments. As a separate study~\cite{loetscher2023assessingrobustnesslidarradar} has shown, LiDAR performance drops to 33\% while depth cameras maintain full-ranging capabilities in adverse conditions. Given these findings, our work diverges from \mbox{TinyLidarNet} by adopting an RGB-D-based supervised learning approach, aiming to benefit from stereoscopic camera sensing's robustness and resilience to lighting variations. Moreover, to provide a fair comparison and to validate our sensor choice, we also trained and deployed \mbox{TinyLidarNet} under the same experimental conditions as our RGB-D models. This allowed us to directly contrast the two sensing modalities and demonstrate the robustness of RGB-D in scenarios with reflective surfaces.

Collectively, these studies highlight the shift in autonomous systems toward multimodal learning and compact neural controllers, yet real-world deployment remains limited. Our goal is to enable autonomous agents to navigate their environment reliably on low-cost, resource-constrained hardware. By deploying efficient RGB-D-based models, we ensure practicality and robustness in real-world scenarios.
% Collectively, these studies highlight the shift toward multimodal learning, recurrent architectures, and biologically inspired neural controllers in autonomous systems. However, real-world applicability and robustness remain largely unexplored. Our goal is to enable autonomous agents to navigate their environment using low-cost, resource-constrained hardware while maintaining reliability and performance. To this end, we focus on deploying compact neural networks with RGB-D input, ensuring both efficiency and practicality in real-world scenarios.

\section{METHODS}\label{sec:methods}
In this section, we present the methodology used to develop and evaluate our neural architectures. We start by describing the hardware setup and our data collection process, followed by the dataset creation pipeline, including the measures implemented to ensure high-quality training data.
\subsection{Hardware Setup and Data Collection}
% hardware setup
The platform we used in this paper is a 1/10-scaled car, referred to as \textit{roboracer} in the autonomous driving research community. Further details about the platform, including in-depth hardware description and state-of-the-art applications are detailed by the platform creators in~\cite{f1tenthcourse}.

The hardware components of the \textit{roboracer} platform include lower-level and upper-level chassis, hosting autonomy elements. The embedded computing platform is a 13th Gen Intel(R) Core(TM) i7-1360P. For optical vision, we integrated Intel RealSense D435i into the hardware stack. According to the datasheet, key properties of the camera include a range of 10m (best performance between 0.3m and 3m), and 69$\degree$$\times$42$\degree$ RGB FOV, 87$\degree$$\times$58$\degree$ Depth FOV. We selected the D435i because it is an active-stereo RGB-D camera with reliable ROS support, combining cost-efficiency and a suitable FOV with robustness to both indoor and outdoor lighting conditions, complementing RGB visual cues and avoiding some of the limitations of LiDAR, such as reduced effective range under bright sunlight and unreliable returns on highly reflective surfaces.

Given the 30fps limitation of the RGB camera, the same number of fps was used for the depth stream. Both camera streams were configured to 848$\times$480px resolution, which is the optimal recommended resolution for the depth camera.

We set up the driving area indoors, using standard \textit{roboracer} pipes to build different track configurations. Our choice was motivated by the pipes typically used during competitions, which provide an excellent application for using a multimodal optical stream to control the steering.

\begin{figure}[bt!]
  \centering
  \begin{subfigure}{0.15\columnwidth}
    \includegraphics[height=2.00cm]{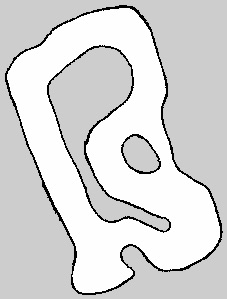}
    \caption{}
    \label{fig:map1}
  \end{subfigure}\hfill
  \begin{subfigure}{0.15\columnwidth}
    \includegraphics[height=2.00cm]{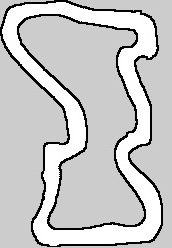}
    \caption{}
    \label{fig:map2}
  \end{subfigure}\hfill
  \begin{subfigure}{0.15\columnwidth}
    \includegraphics[height=2.00cm]{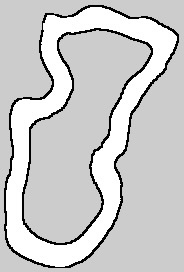}
    \caption{}
    \label{fig:map3}
  \end{subfigure}\hfill
  \begin{subfigure}{0.15\columnwidth}
    \includegraphics[height=2.00cm]{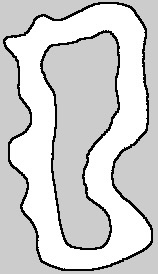}
    \caption{}
    \label{fig:map4}
  \end{subfigure}\hfill
  \begin{subfigure}{0.15\columnwidth}
    \includegraphics[height=2.00cm]{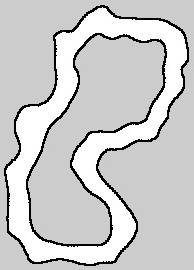}
    \caption{}
    \label{fig:map5}
  \end{subfigure}\hfill
  \caption{Tracks built for human expert data collection.}
  \label{fig:maps_dataset}
\end{figure}

We built a series of 5 tracks with a variation of successive turns, of different curvatures and lengths, as seen in Figure~\ref{fig:maps_dataset}. In order to increase the dataset size, all tracks were driven both clockwise and anti-clockwise.

A driver with 10 years of driving experience was chosen to control the car. The driver was given control of the vehicle through a racing steering wheel and pedals. We found that due to the narrow FOV of the camera, the driver was most comfortable driving with a maximum speed of 0.9m/s.

The data flow of the driving stack during data collection is illustrated in Figure~\ref{fig:stackdiagram}. To simulate a first-person point of view, we redirected the RGB stream from the camera to a second device with a monitor placed in front of the driver. While driving, rosbags were being recorded on all available car topics, excluding the RGB topic, since the RGB data stream was being saved as video files. 

\begin{figure}[bt!]
    \centering
    \includegraphics[width=0.9\linewidth]{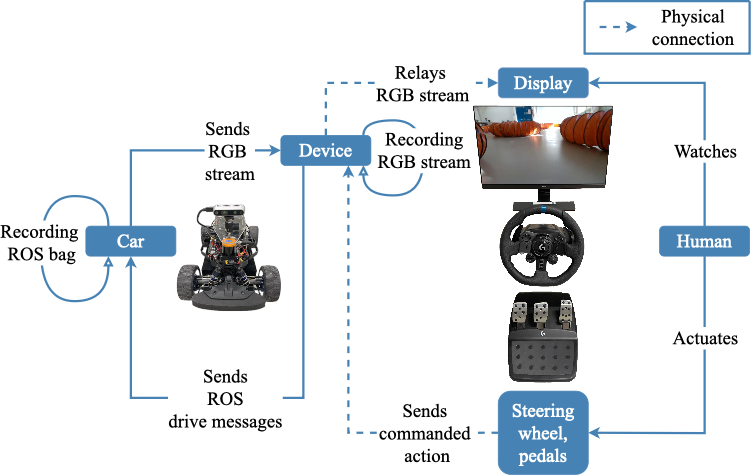}
    \caption{Driving stack overview for data collection step.}
    \label{fig:stackdiagram}
    \vspace{-4ex}
\end{figure}

% The driver was allowed a few trial runs on each track before the data recording was started in order to get accustomed to the camera's field of view and the difficulty of the track. For Figure~\ref{fig:map1}, we instructed the driver to take the longer route at the intersection.

\begin{figure*}[t!]
    \centering
\includegraphics[width=0.9\linewidth]{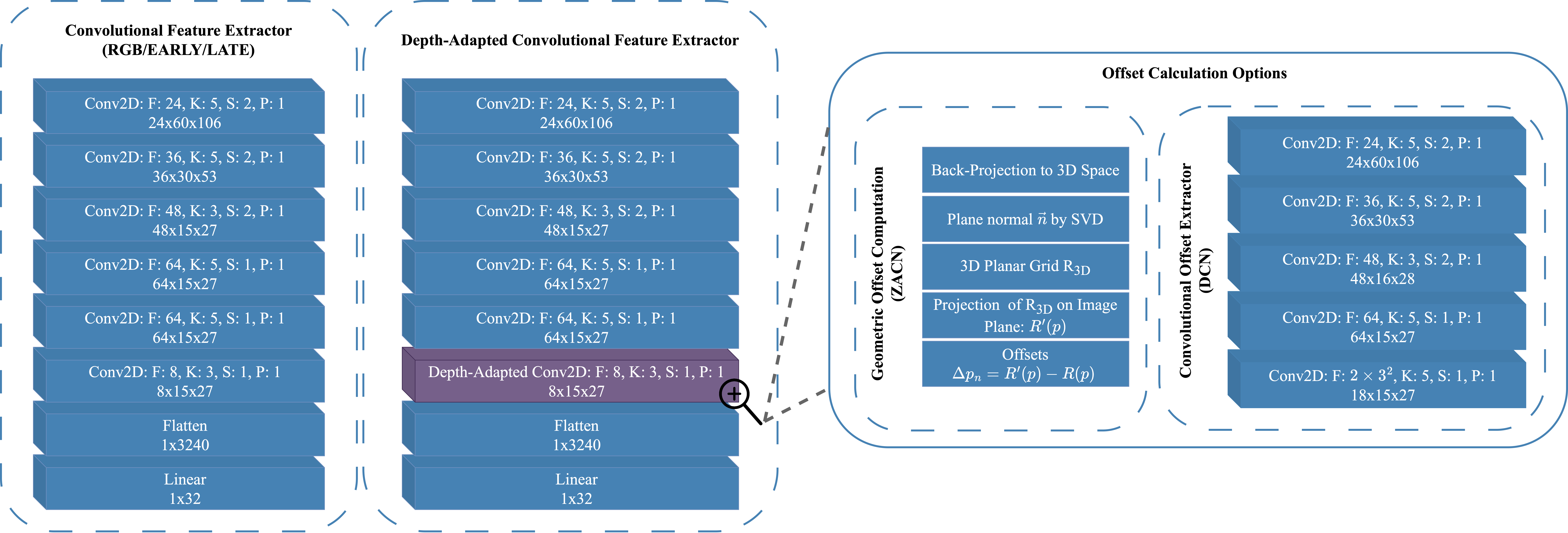}
    \caption{The sequence of layers in the convolutional heads is as follows. The Convolutional Feature Extractor is used with an RGB input of size 3×120×212. In the early fusion (EARLY) method, it processes an input of size 4×120×212. In the late fusion (LATE) method, it is used twice, treating RGB and depth as two separate input streams of sizes 3×120×212 and 1×120×212, respectively, and concatenating their features after the last layer. The Depth-Adapted Convolutional Feature Extractor has two versions: (1) the DCN, which includes a Convolutional Offset Extractor, and (2) the ZACN~\cite{wu2020depthadaptedcnnrgbdcameras}, which incorporates Geometric Offset Computation. In the hyperparameters section: F denotes the number of filters, K the kernel size, S the stride, and P the padding.}
    \label{fig:microview}
    \vspace{-4ex}
\end{figure*}

\subsection{Dataset Collected for Open-Loop Training}\label{sec:dataset}
% dateset info

Firstly, the messages on the depth topic were extracted from each rosbag. Their content consists of uint16 1-channel images representing the measured distance for each pixel. We parsed each frame to RGB by creating a colormap on the pixels based on their distance with respect to the maximum measured value. Then, we generated depth videos from each rosbag and stamped every frame with its index.

As a second step, we manually synchronized the original videos with the depth from the rosbags. The original videos were broken down and stamped by frame number. Before the start of each recording session, we presented visual cues to the camera to help with synchronization. Furthermore, we logged frame numbers corresponding to the start and end of each lap in the recordings.

Thirdly, the resolution of the streams was downscaled to 1/4th of the initial scale. For this purpose, we tested several rescaling algorithms: nearest neighbor, Gaussian, Lanczos~\cite{lanczos1950iteration}, and Catmull-Rom~\cite{catmull1974class}. As expected, nearest neighbor downsampling produced grainier results, while Gaussian downsampling blurred the stream. As neither was desirable, Lanczos-3 was chosen over Catmull-Rom for producing overall sharper results.

Lastly, the depth topic messages were iterated over based on the lap frame intervals in each recording. With the timestamp of each frame message, we then found the corresponding steering commands as in~\cite{resch2023attention} by interpolating the values between two timestamps. In this way, each RGB frame and depth message was labeled with the commanded steering angle and saved as \texttt{.npz} files. We obtained a dataset of approximately $100\,000$ labeled frames. % from all collected data.

% \subsection{Dataset Analysis}

% \begin{figure}[hbt!]
%     \centering
%     \includegraphics[width=\linewidth]{graphics/graphs/boxplot_steering_angle_distribution_by_map.png}
%     \caption{Distributions of recorded steering angles in 1 lap on each track. Negative values correspond to steering right, whereas positive values to steering left.}
%     \label{fig:steeringboxplot}
% \end{figure}

% Through this approach, a dataset of approximately $100\,000$ labeled frames was obtained from all collected data. Figure~\ref{fig:steeringboxplot} offers an insight into not only the variety of commands reached on driven tracks but also the difference in distribution between driving directions. 

% Naturally, anti-clockwise driving directions keep strictly positive median values, which correspond to the driver keeping left while driving. In contrast, clockwise driving is characterized by negative median values. 

% Moreover, the spread of the distributions matched the levels of navigation difficulty previously implied: narrow distributions (tracks 1 and 4) were also faster to navigate, whereas the broadest distribution (track 2) registered the longest navigation times. 

% Lastly, the outliers on the \gls{ACW} direction on tracks 2, 3, and 5 correspond to very sharp right turns.

\subsection{Work Environment}
The machine used for training the models runs Ubuntu 20.04.6 and has $3\,072$ CUDA cores and 12 GB of GDDR5 memory. We chose PyTorch due to its out-of-the-box interoperability and extensive computer vision package support. %, and configured a Conda virtual environment. %Code is available in our \href{https://github.com/clementlarisa/multimodal-rgbd-agents}{GitHub repository}. %Not for the double blind

%Several measures can be taken to ensure reproducibility, although identical behavior cannot be guaranteed between all different runs. With  this in mind, all sources of non-determinism that are adjustable through the PyTorch framework were investigated and seeded with a chosen value. %M: it's too detailed

\section{EXPERIMENTAL SETUP}\label{sec:model_architectures}

%Starting from the approach of the authors of~\cite{Lechner_Hasani_Amini_Henzinger_Rus_Grosu_2020}, 
In total, 20 model architectures were designed and implemented. 
We focused on one unimodal and 4 multimodal approaches for feature extraction.
%Four of the feature extractor architectures diverge from~\cite{Lechner_Hasani_Amini_Henzinger_Rus_Grosu_2020} because of the integration of multimodal input and one architecture using the unimodal RGB emulates the exact original method. 

\subsection{Unimodal RGB Setup}
The implemented unimodal architecture follows~\cite{Lechner_Hasani_Amini_Henzinger_Rus_Grosu_2020}, with a convolutional head illustrated in Figure~\ref{fig:microview} which feeds the output to an \gls{RNN} of choice. This aims to extract relevant features of the RGB input and use them as input to \gls{RNN}s to make memory-based steering commands. We refer to this method as RGB in the next sections.

\subsection{Multimodal RGB-D Setups}
(i) The first multimodal approach is an early fusion approach (referred to as EARLY), introducing the depth values as a fourth channel to the RGB input of the previously described convolutional head. This type of fusion achieved the best results in~\cite{Mújica-Vargas2024} and~\cite{MultimodalEnd-to-EndAutonomousDriving}. Our specific architecture is shown in Figure~\ref{fig:microview} as the Convolutional Feature Extractor.

(ii) The second multimodal method is a late fusion method (referred to as LATE). This essentially presents a two-stream architecture, where different \gls{CNN}'s are used for each stream. 
% However, while the first approach finds ways to transform the 1-channel depth input into 3-channel RGB input, this paper immediately uses the original 1-channel values. The motivation behind this is that the former uses the same pre-trained \gls{CNN} for each stream (i.e. ImageNet), which was trained on RGB inputs. This is not necessary when the models are trained from the ground up.

(iii) The final method illustrated in Figure~\ref{fig:microview} as Depth-Adapted Convolutional Feature Extractor, is processing the multimodal input using Deformable \gls{CNN}. As the goal of deformable convolutions is to use spatial adaptation of kernels, we only use the operation once, in the last layer. In this way, the most relevant features extracted from RGB are simply enhanced by offsets based on depth information. To this end, two techniques were utilized for calculating the convolutional offsets for the deformable layer.

(iii-a) The first technique in this subcategory relies on extracting the offsets from the single-channel depth data using \gls{CNN}. We refer to this as DCN.

(iii-b) The second technique, which we refer to as ZACN, is based on~\cite{wu2020depthadaptedcnnrgbdcameras}. The intrinsic camera parameters used by this method were obtained from the data collection device. % \ref{sec:data}.

\subsection{Recurrent Models}
The models are further differentiated from each other based on 4 \gls{RNN} backbone choices: \gls{LSTM}, \gls{LTC}, \gls{CfC} and 
\gls{LRC}. The \gls{LTC} and \gls{CfC} backbones use a \gls{NCP} of 19 neurons, i.e., 12 inter-neurons, 6 command neurons, and 1 motor neuron, as LTCs were configured in~\cite{Lechner_Hasani_Amini_Henzinger_Rus_Grosu_2020}. We adapted CfCs to this setup as they are the closed-form solutions of LTCs. The \gls{LSTM} and LRC architectures are configured to have a standard size of 64 features in the hidden state. All \gls{RNN}'s use a sequence length of 16 frames. At 30 fps, 16 frames correspond to $\approx$0.5 seconds of temporal context. This value balances capturing sufficient motion dynamics (turn anticipation) and staying within memory constraints.

\subsection{Training and Metrics}\label{sec:training_metrics}
Regarding the dataset, the order of the 10 recordings (2 directions for each track) was shuffled. Starting from this randomized order, we split the dataset into a 60/20/20 ratio representing training, validation, and test data. The split datasets are further shuffled before being fed to the models.

All run configurations include arguments for the \gls{RNN} model, input mode, the type of per-channel normalization in the normalization layer, padding, and learning rates. We experimented with Min-Max normalization, Z-Score normalization, and Min-Max followed by Z-Score. During training, we set the maximum epoch number to 100 epochs with a batch size of 20, constrained by the available computational resources. To prevent overfitting, training stops early if the validation loss does not improve for three consecutive epochs. We used Tensorboard to log and analyze all models' training, validation, and testing results.

We used the Adam optimization algorithm ~\cite{kingma2014adam} with learning rates of $\{10^{-3}, 10^{-4}, 10^{-5}\}$ and $\epsilon=10^{-7}$ for numerical stability. To predict continuous steering commands from multi-dimensional input, we used the \gls{MSE} as loss function, where values closer to 0 indicate more accurate estimations.

\section{RESULTS}\label{sec:results}
In this section, we summarize our findings on the various model architectures presented in Section~\ref{sec:model_architectures}. First, we describe the results from the passive test using the collected dataset, followed by deployment results of the best models.

\subsection{Open-loop Results}
We follow the training procedure described in Section~\ref{sec:training_metrics}. Since we use supervised learning at each timestep, the next input frame is not influenced by the model’s previous decision, which is known as an open-loop setting.
% We started by training the unimodal RGB models with all possible configurations. We then anaylzed the validation losses of each model, and chose the best normalization technique, learning rate and padding of each RNN architecture. We proceeded to train the multimodal models using these setups.

Table~\ref{tab:dataset_test} presents our best open-loop test loss results using the recorded dataset. We found that for each feature extractor, the two best-performing models are LSTM and LRC, except for RGB, where CfC performed better than LRC.

\begin{table}[tb]
    \centering
    \caption{Test results, averaged over 3 seeds. Models were trained, validated, and tested on the collected dataset. Loss values are scaled by \( \times 10^{-3} \). The two best-performing recurrent models for each feature extractor are highlighted.}
    \begin{tabular}{lccccc}
        \toprule
        & LSTM & LTC & CfC & LRC \\
        \midrule
           EARLY &  $\mathbf{7.53 \pm 0.49}$  &   $ 11.73 \pm 3.82 $ & $ 12.60 \pm 4.42 $ & $\mathbf{8.23 \pm 1.71}$  \\
           LATE & $\mathbf{5.93 \pm 0.77}$   &  $ 9.93 \pm 1.02 $  & $ 8.97 \pm 1.63 $  & $\mathbf{7.80 \pm 0.92}$\\
           ZACN & $\mathbf{8.27 \pm 1.22}$    &  $ 10.27 \pm 0.90 $   & $ 9.53 \pm 1.17 $  & $\mathbf{8.07 \pm 1.10}$ \\
           DCN &   $\mathbf{8.77 \pm 1.63}$ & $ 22.50 \pm 9.75 $ & $ 11.65 \pm 1.85 $ & $\mathbf{9.37 \pm 1.88}$ \\
           RGB & $\mathbf{9.57 \pm 2.37}$ &  $ 13.30 \pm 1.91 $  &  $\mathbf{11.87 \pm 1.74}$ & $ 12.00 \pm 2.14 $ \\
        \bottomrule
    \end{tabular}
    \label{tab:dataset_test}
    \vspace{-3ex}
\end{table}

\subsection{Closed-loop Results}
After training and evaluating the models on the passive dataset, we were interested in \textit{how they would adapt to a closed-loop setting}, i.e., when the model's steering predictions are fed back into the control loop of the vehicle and directly influence subsequent observations. Since we had a total of 20 models (5 feature extractors × 4 RNN options), we chose to first test only the most promising ones that demonstrated good performance on the dataset. To ensure a balanced evaluation of both the feature extractors and the RNNs, we created test groups based on both components. Specifically, we selected the best two recurrent models for each feature extractor and the best two feature extractors for each recurrent model across all seeds. %The results are presented in Table~\ref{tab:cnn_group}.

\begin{figure}[bt!]
\centering
\includegraphics[width=0.8\columnwidth]{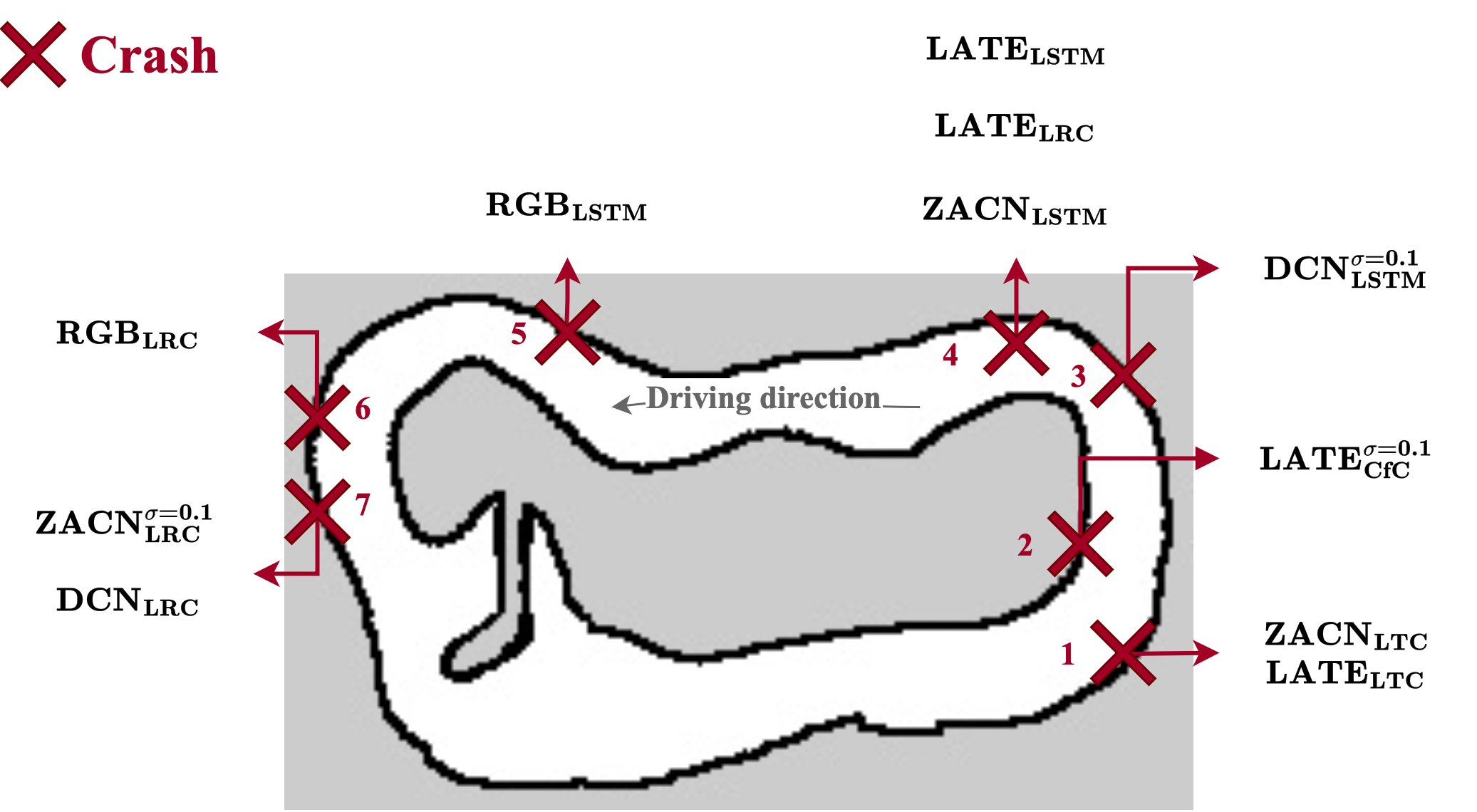}
\caption{Extraction of a map used during closed-loop active testing. Problematic turns are marked with an 'X' to indicate crashes. Labels specify which model crashed at each point.}
\label{fig:mapactive}
\vspace{-3ex}
\end{figure}

We first found that a closed-loop setup is very challenging for many models, with only a subset completing 5 continuous laps without crashing on Map~\ref{fig:mapactive}. No RGB-based models were able to complete this task. We marked the sections in Figure~\ref{fig:mapactive} where crashes occurred. Interestingly, the LATE feature extractor only succeeded with the CfC model, even though it was not among the two best recurrent models for the LATE type in the passive open-loop test (as shown in Table~\ref{tab:dataset_test}). However, we evaluated it because CfCs demonstrated the best performance with this feature extractor method, which highlights the effectiveness of our double grouping approach for feature extractors and recurrent models, respectively. This also illustrates the sim-to-real gap, highlighting the performance mismatch between the open-loop setup and real hardware deployment.

\begin{figure}[bt!]
    \centering
    \includegraphics[width=0.99\linewidth]{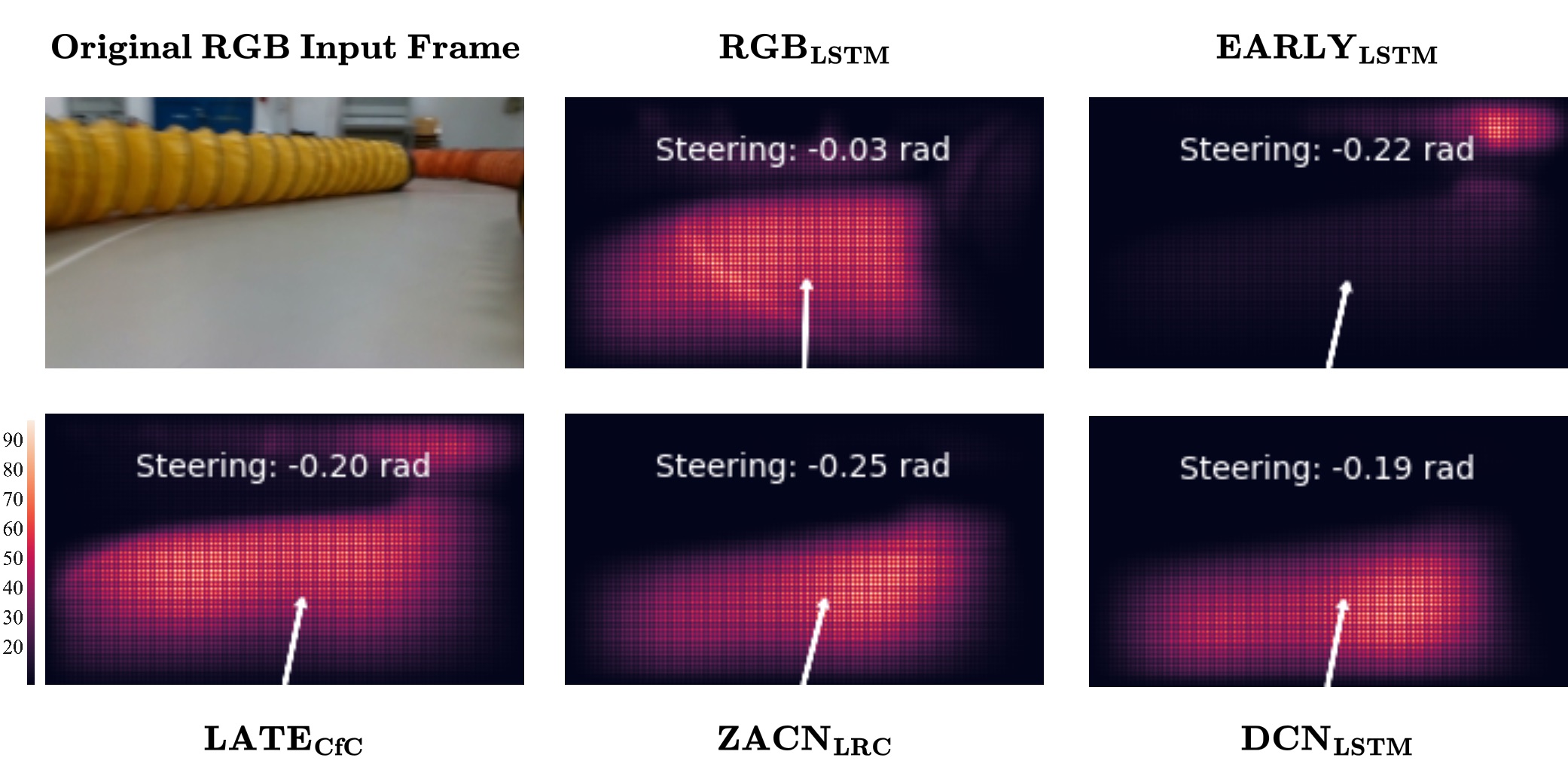}
    \caption{Attention maps during closed-loop testing, evaluated on RGB\textsubscript{LSTM}, and the models that achieved full autonomy for 5 consecutive laps. The chosen frame is prior to the RGB\textsubscript{LSTM} model crash at landmark 5 of Figure~\ref{fig:mapactive}.}
    \label{fig:allnetworksattention}
\end{figure}

We explored the underlying attention mechanism of these models to gain insight into which parts of the input they focus on the most. To calculate this, we used the VisualBackProp method~\cite{visualbackprop}, where bright regions indicate where the model focused its attention. Attention maps from Map~\ref{fig:mapactive} are displayed in Figure~\ref{fig:allnetworksattention}. We found that the EARLY\textsubscript{LSTM} maintains its focus on the more distant regions, while the other models focus more on the immediate road. The attention of the LATE\textsubscript{CfC}, ZACN\textsubscript{LRC}, and DCN\textsubscript{LSTM} models tends to explore the right side of the road more, while the RGB\textsubscript{LSTM} model keeps its attention evenly in front, without focusing on the turn. We hypothesize that a similar lack of focus on the turn contributed to the RGB\textsubscript{LSTM}'s crash.

\subsection{How Robust Are the Deployed Models?}
We selected models that completed 5 consecutive laps on the track without crashing for further evaluation under noisy conditions. These models include EARLY\textsubscript{LSTM}, LATE\textsubscript{CfC}, ZACN\textsubscript{LRC}, and DCN\textsubscript{LSTM}. To test their robustness, we added additional Gaussian noise with a mean of 0 and a variance of 0.1 to the input stream. Only the EARLY\textsubscript{LSTM} model handled successfully this challenging condition, while the others failed, resulting in crashes at Points 2, 7, and 3, as shown in Figure~\ref{fig:mapactive}, respectively. We hypothesize that this is due to the combination of the smart early fusion mechanism and the optimized LSTM implementation in this resource-constrained setup.

% \begin{figure}[bt!]
%     \centering
%     \includegraphics[width=0.75\linewidth]{graphics/graphs/ssim_noise_01_crash.png}
%     \caption{The Structural Similarity Index (SSIM) values for the EARLY\textsubscript{LSTM}, LATE\textsubscript{CfC}, ZACN\textsubscript{LRC}, and DCN\textsubscript{LSTM} models measure the networks' attention robustness under noise. EARLY\textsubscript{LSTM} performed the best, with values close to 1, indicating stronger resilience against noisy input than the rest of the models analysed.}
%     \label{fig:ssimcrash}
% \end{figure}

\begin{figure}[bt]
    \centering
    \includegraphics[width=0.99\linewidth]{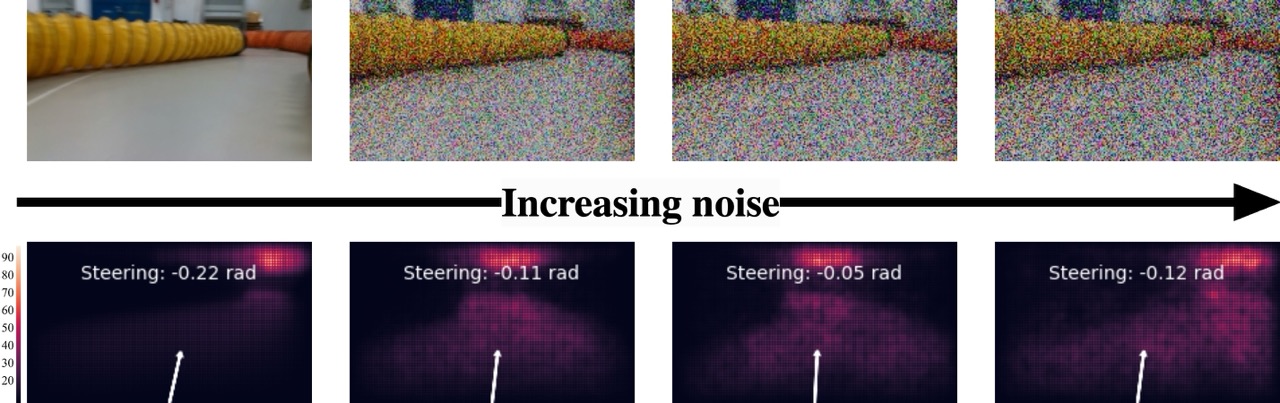}
     % \vspace{0.5em} % <-- this should add space but doesn't work
    \includegraphics[width=0.7\linewidth]{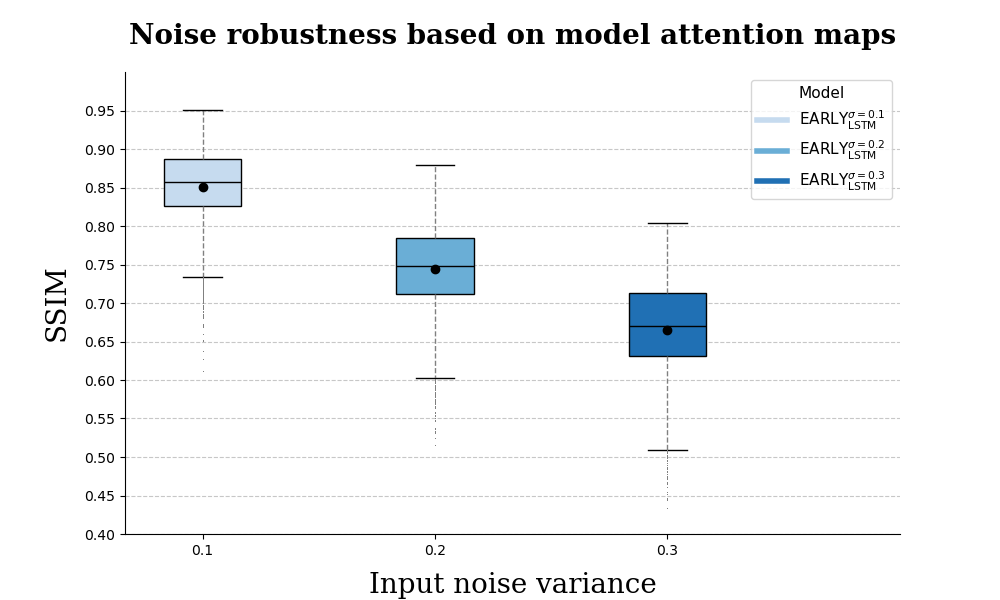}
    \caption{Closed-loop testing with no-noise and increasing Gaussian noise with variance of $\sigma=0.1$, $0.2$ and $0.3$ on the EARLY\textsubscript{LSTM} model. 
    %Evaluated on the EARLY\textsubscript{LSTM} model, as it was the best-performing model being able to receive additional increasing input noise. 
    Top: Attention maps demonstrate that the model could consistently maintain its attention on the same region. Bottom: High SSIM index for attention maps between no-noise and increased Gaussian noise levels.}
    \label{fig:earlylstmattention}
\end{figure}

% %How the test setup differed from the training data
% \begin{figure}[bt!]
%     \centering
%     \includegraphics[width=0.75\linewidth]{graphics/graphs/ssim_early_fusion.png}
%     \caption{Robustness against noise during closed-loop testing, evaluated on the EARLY\textsubscript{LSTM} model, which managed to maintain 100\% autonomy without crashing, under increased noise. Here, we illustrate the SSIM index for attention maps between no-noise and increased Gaussian noise with variance of $\sigma=0.1$, $0.2$ and $0.3$, respectively. }
%     \label{fig:ssimearly}
%     % \vspace{-3ex}
% \end{figure}

%To analyze this quantitatively, we want to see how much the attention of these models gets distorted when extra noise is introduced. For this, we used the Structural Similarity Index (SSIM)~\cite{SSIM}, which assigns a score between 0 and 1 to measure the similarity between two images. In our case, these images represent pairwise comparisons of attention maps under noise-free and noisy conditions. SSIM values close to 1 indicate that the network's attention remained unchanged, demonstrating its resilience. The SSIM values between the no-noise and $\sigma=0.1$-variance noise for these models are shown in Figure~\ref{fig:ssimcrash}. The significantly higher SSIM value of the EARLY\textsubscript{LSTM} supports our finding that it can tackle noise with more robust attention, while the other models failed under this condition.

To further assess the robustness of the promising EARLY\textsubscript{LSTM} model, we gradually increased the noise level to 0.2 and then to 0.3. Impressively, the model remained robust even at these high noise levels. We were particularly interested in how this resilience relates to the model's attention maps - specifically, how well it maintains focus on key input features without being distracted by noise during steering. This is shown at the top of Figure~\ref{fig:earlylstmattention}. % To calculate the attention maps, we used the VisualBackProp method~\cite{visualbackprop}, where bright regions indicate where the model focused its attention.

To analyze this quantitatively, we want to see how much the attention of this model gets distorted when extra noise is introduced. For this, we used the Structural Similarity Index (SSIM)~\cite{SSIM}, which assigns a score between 0 and 1 to measure the similarity between two images. In our case, these images represent pairwise comparisons of attention maps under noise-free and noisy conditions. SSIM values close to 1 indicate that the network's attention remained unchanged, demonstrating its resilience. This are presented at the bottom of Figure~\ref{fig:earlylstmattention}, which demonstrates that even with a high level of noise, the EARLY\textsubscript{LSTM} managed to keep its attention similarity close to 1, with $86\%, 75\%$ and $67\%$, respectively. Although there is a decrease in SSIM values as noise increases, the rate of decrease is not linear: it slows down progressively, with values decreasing by $-14\%$, $-11\%$, and $-8\%$, respectively. This suggests that while the model is affected by noise, it retains much of its attention stability even at higher noise levels. This resilience could be a key factor in its consistent performance under these challenging, noisy conditions.

% \begin{figure}[hbt!]
% \centering
% \includegraphics[width=\columnwidth]{graphics/dome/human_1lap.png}
% \caption{Trajectory of the human driver in of the test track via object tracking.}
% \label{fig:domehuman}
% \end{figure}
% \begin{figure}[bt!]
% \centering
% \includegraphics[width=\columnwidth]{graphics/dome/model_1lap.png}
% \caption{Trajectory of our best model, the Early\textsubscript{LSTM}, of the test track via object tracking. The model remained 100\% autonomous even with increased noise level added to the RGB-D stream input.}
% \label{fig:domemodel}
% \end{figure}

% Use in your thesis
% \begin{table}[h]
%     \centering
%     \renewcommand{\arraystretch}{1.2} 
%     \begin{tabular}{lcccc}
%         \toprule
%         \textbf{Model} & \textbf{$\sigma$ = 0.0} & \textbf{$\sigma$ = 0.1} & \textbf{$\sigma$= 0.2} & \textbf{$\sigma$ = 0.3} \\
%         \midrule
%         EARLY\textsubscript{LSTM} & 8.267 ms & 8.693 ms & 7.205 ms & 4.986 ms \\
%         LATE\textsubscript{CfC} & 13.854 ms & 9.778 ms & \xmark & \xmark \\
%         ZACN\textsubscript{LRC} & 20.549 ms  & 14.338 ms & \xmark & \xmark \\
%         DCN\textsubscript{LSTM} & 18.552 ms & 15.191 ms & \xmark & \xmark \\
%         \bottomrule
%     \end{tabular}
%     \caption{Average Latency (ms) of models who autonomously drove 5 consecutive laps when $\sigma = 0.0$. The average throughput for all models was $\approx20fps$}
%     \label{tab:latency}
% \end{table}

\subsection{Out-of-Distribution Testing}

\begin{figure}[bt!]
    \centering
    \includegraphics[width=0.85\linewidth]{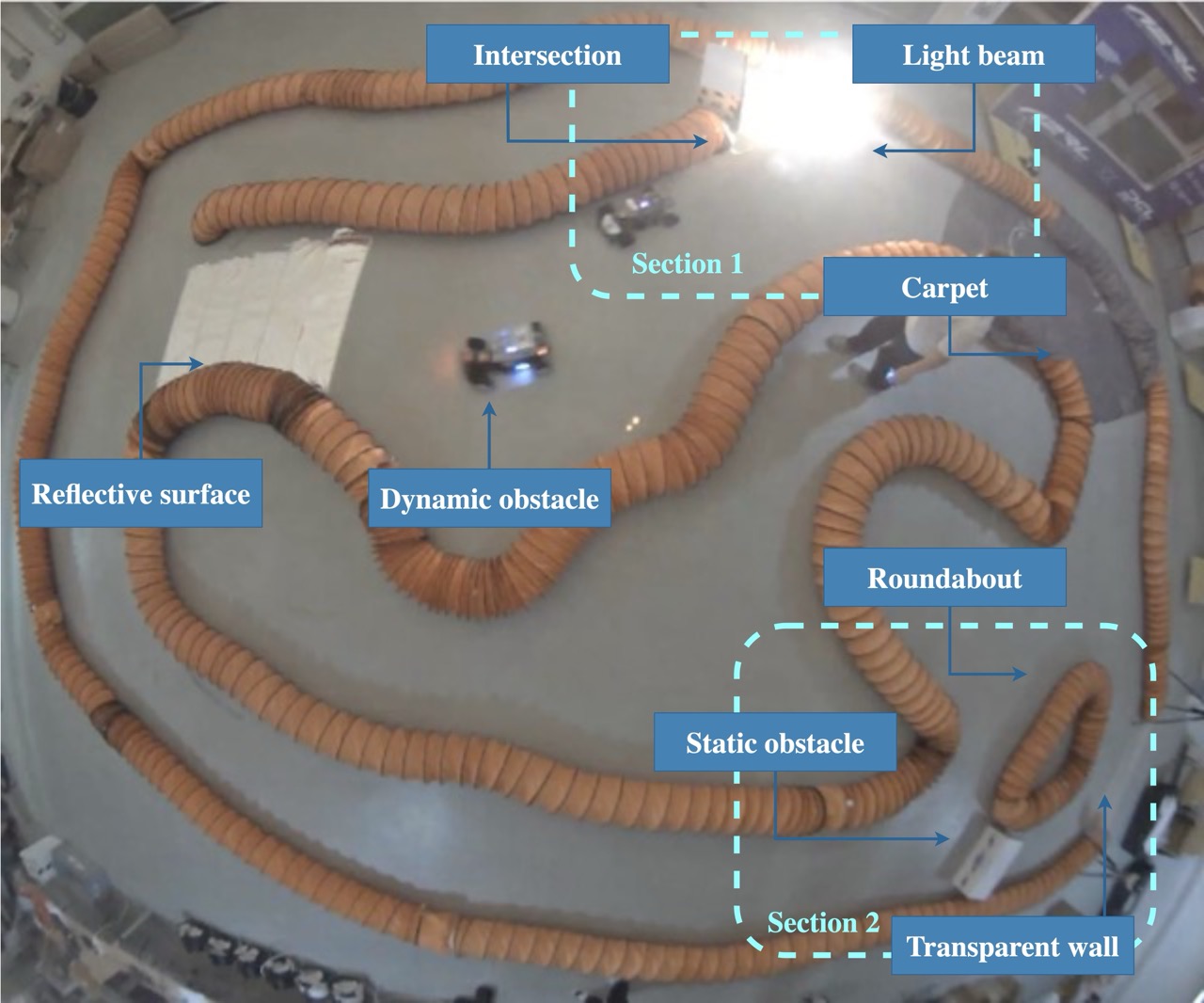}
    \caption{Labeled testing circuit in dark conditions. The figure shows one of the testing tracks with labeled challenges. Different variations were conducted by moving the static obstacles and light sources, repositioning reflective foil on different walls (turns, roundabout), introducing gaps by removing pipe segments or adding/removing the transparent wall (plastic wall).}
    \label{fig:full-track}
\end{figure}

To understand EARLY\textsubscript{LSTM}, we constructed a configurable testing circuit with multiple challenging elements (Fig.~\ref{fig:full-track}). The environment included reflective surfaces, directed light beams, carpet material, transparent walls, decision-making points: intersection, roundabout, and both static and dynamic obstacles (a second \textit{roboracer}, controlled by a human). These conditions were not fixed: obstacles and light sources were moved between runs, reflective foil was attached to different wall segments, gaps were opened by removing pipes, and new obstacles were introduced while the vehicle was driving. This ensured that the model was exposed to a wide distribution of untrained situations, testing the ability to generalize in closed-loop deployment.

\begin{figure}[bt!]
    \centering
    \includegraphics[width=0.55\linewidth]{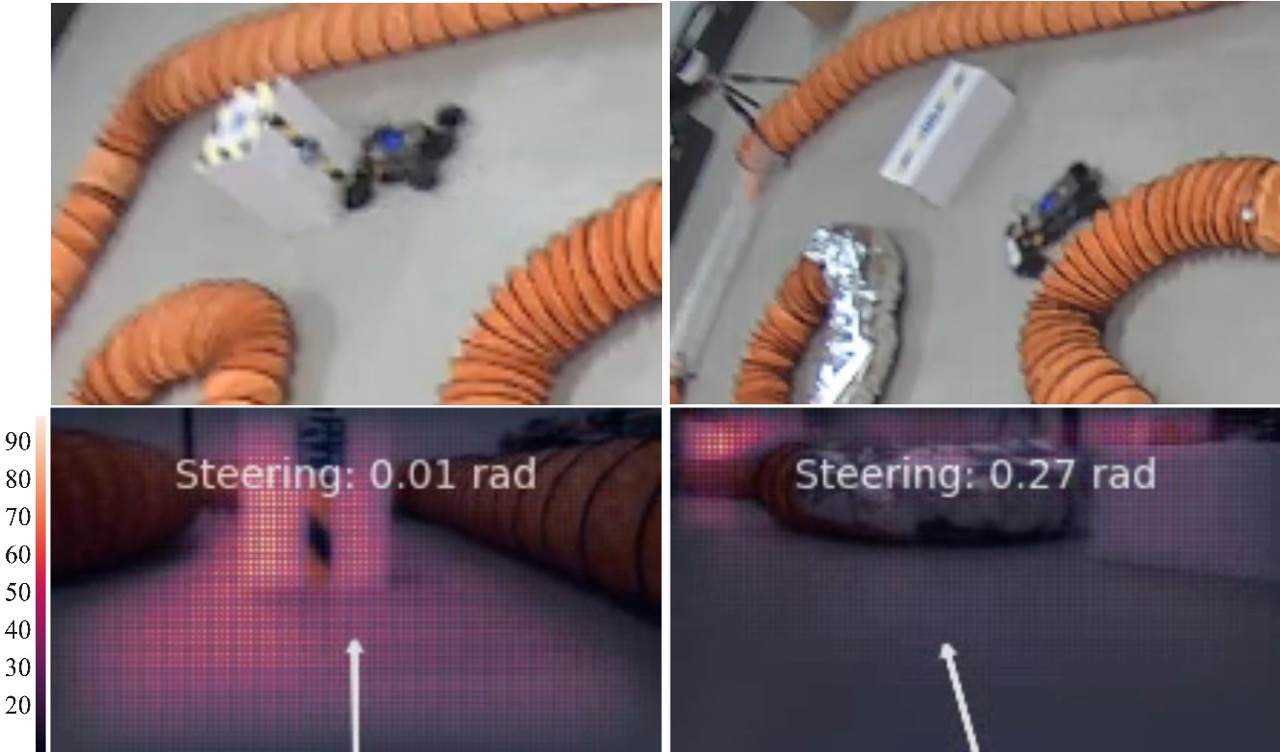}
    \caption{Comparison of RGB\textsubscript{LSTM} and EARLY\textsubscript{LSTM} models in closed-loop obstacle interaction. Overlaid attention maps highlight the difference: while the RGB model remains focused on the blocked route (left), the depth-based EARLY\textsubscript{LSTM} shifts its attention toward the free path (right).}
    \label{fig:rgb-obstacle}
    % \vspace{-3ex}
\end{figure}
One of the clearest differences between modalities emerged during obstacle tests. As shown in Fig.~\ref{fig:rgb-obstacle}, EARLY\textsubscript{LSTM} successfully avoided an unseen obstacle in the roundabout by immediately redirecting its trajectory toward the free branch. In contrast, the RGB\textsubscript{LSTM} failed to perceive the obstruction, continued to predict a near-straight steering command, and collided with the object. Notably, neither model was trained for obstacle avoidance, yet depth provided an implicit geometric cue that guided safe behavior, while RGB-only perception lacked the necessary information.

% \begin{figure}[bt!]
%     \centering
%     \includegraphics[width=0.75\linewidth]{graphics/new_experiments/intersection-choice.png}
%     \caption{Closed-loop behavior at intersections: EARLY\textsubscript{LSTM} model took different paths - left or right - depending on its approach, as shown by overlaid attention maps.}
%     \label{fig:intersection_behavior}
% \end{figure}

\begin{figure}[bt!]
    \centering
    \begin{subfigure}[t]{0.39\linewidth}
        \centering
        \includegraphics[height=0.9cm]{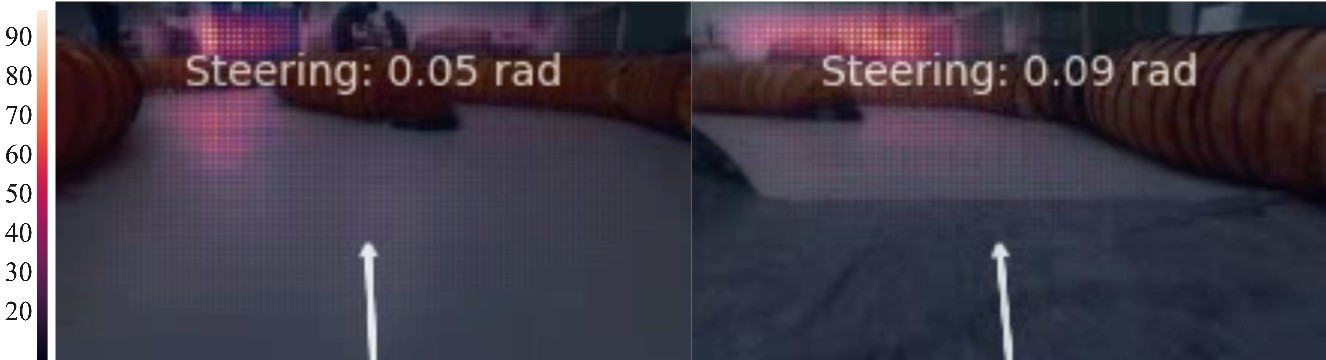}
        \caption{}
        \label{fig:intersection_behavior}
    \end{subfigure}
    \hfill   
    \begin{subfigure}[t]{0.59\linewidth}
        \centering
        \includegraphics[height=0.9cm]{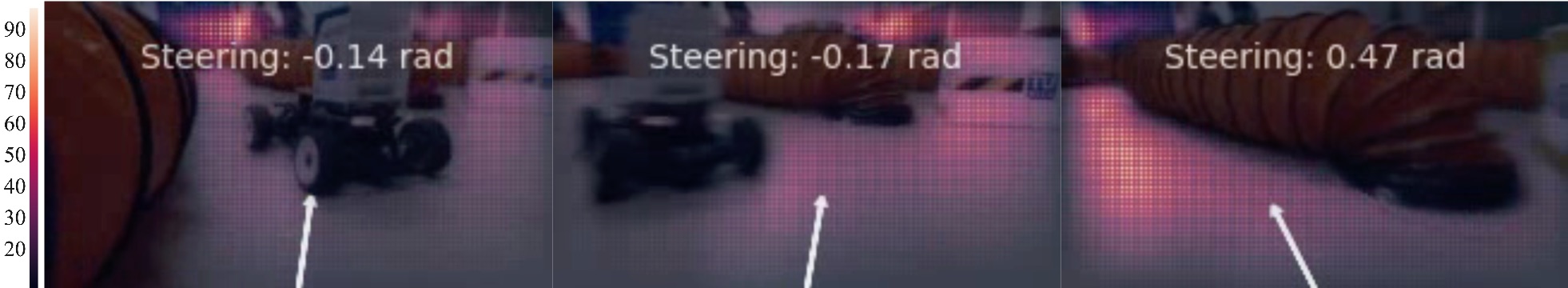}
        \caption{}
        \label{fig:car7-intersection}
    \end{subfigure}
    
    \caption{Behavior of the EARLY\textsubscript{LSTM} model at intersections, illustrated with overlaid attention maps. (a) Taking left or right paths depending on its approach. (b) Handling dynamic (other car) and static (white box) obstacles.}
    \label{fig:combined_intersections}
    \vspace{-3ex}
\end{figure}

Although our training data did not contain intersections or explicit decision-making commands, we observed that the RGB-D models were able to navigate such scenarios in closed loop. Figure~\ref{fig:intersection_behavior} illustrates two examples where the vehicle approached an intersection from different trajectories. In the first case, the model exited the preceding turn with a shallow angle, allowing visibility of both branches, and ultimately chose the left branch by steering slightly left (0.05 rad). In the second case, the vehicle entered at a sharper angle, giving greater visibility into the right branch, and the model steered accordingly (0.09 rad). This behavior suggests that the controller relied on forward visibility when choosing a path, effectively preferring the branch with more open space ahead. Importantly, this decision-making emerged without explicit supervision for intersections, highlighting the capacity of the RGB-D recurrent models to generalize beyond their training distribution and exploit geometric cues available in the sensory input.

\begin{figure}[bt!]
    \centering
    \includegraphics[width=0.99\linewidth]{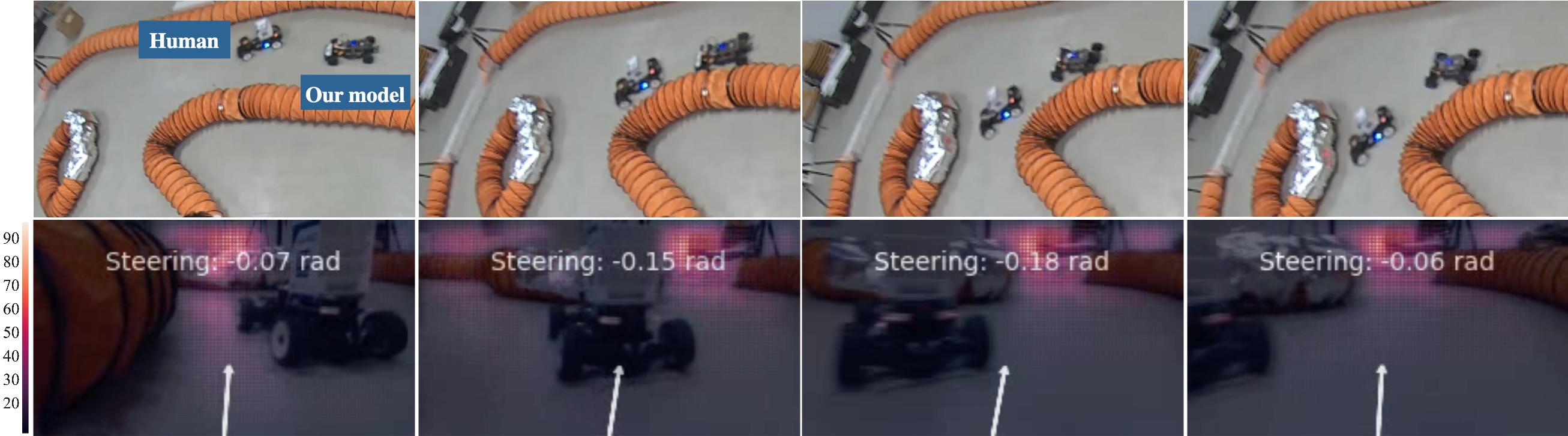}
    \caption{Closed-loop interaction with a dynamic obstacle. Top row: dome-camera view of two \textit{roboracer} vehicles on the circuit (the front vehicle controlled by a human expert). Bottom row: overlaid attention maps from our EARLY\textsubscript{LSTM} model with corresponding overlaid attention maps and steering commands.}
    \label{fig:car7-roundabout}
    \vspace{-2ex}
\end{figure}
Introducing a second, human-driven \textit{roboracer} created dynamic interaction scenarios. Our EARLY\textsubscript{LSTM} handled these cases without explicit training for multi-agent driving. Figure~\ref{fig:car7-roundabout} shows one critical example: as the human-driven vehicle cut in front, the autonomous car adjusted both its attention and steering trajectory in real time. The overlaid attention maps clearly illustrate this adaptation, initially focusing left to overtake, then rapidly shifting right when space became constrained, leading to a smooth avoidance maneuver. This new behavior shows the capacity of our model to exploit RGB-D cues for reactive obstacle avoidance in closed-loop settings.
% \begin{figure}[bt!]
%     \centering
%     \includegraphics[width=0.99\linewidth]{graphics/new_experiments/car7-intersection.png}
%     \caption{Closed-loop behavior of EARLY\textsubscript{LSTM} with dynamic (other car) and static obstacles (white box) at intersection point.}
%     \label{fig:car7-intersection}
% \end{figure}

%Closed-loop behavior at intersections: EARLY\textsubscript{LSTM} model took different paths—left or right—depending on its approach, as shown by overlaid attention maps.
%Closed-loop behavior of EARLY\textsubscript{LSTM} with dynamic (other car) and static obstacles (white box) at intersection point.

% \begin{figure}[bt!]
%     \centering
%     \includegraphics[width=0.75\linewidth]{graphics/new_experiments/intersection-choice.png}
%     \caption{Closed-loop behavior at intersections: EARLY\textsubscript{LSTM} model took different paths - left or right - depending on its approach, as shown by overlaid attention maps.}
%     \label{fig:intersection_behavior}
% \end{figure}

In Figure~\ref{fig:car7-intersection}, a static obstacle blocked one branch of the intersection while the human-driven \textit{roboracer} cuts across the other. In this situation, depth cues indicated no free space to the right, while the human vehicle constrained the left. The model appears to fall back on RGB cues, focusing on the floor texture and up ahead, successfully avoiding both obstacles. This behavior highlights the complementarity of RGB and depth: when depth cues were limited, RGB provided sufficient contextual information to maintain navigation.

\begin{figure}[bt!]
    \centering
    \includegraphics[width=0.6\linewidth]{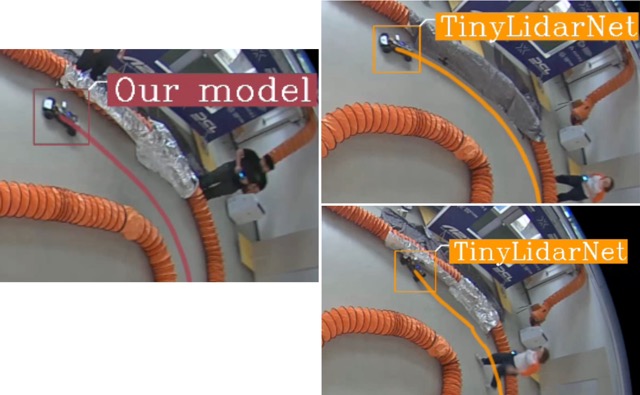}
    \caption{Comparison of EARLY\textsubscript{LSTM} with \mbox{TinyLidarNet} under reflective-surface conditions.}
    \label{fig:tln}
    \vspace{-3ex}
\end{figure}

We deployed \mbox{TinyLidarNet} for comparison on our test track. Under normal conditions, \mbox{TinyLidarNet} performed reliably, producing smooth trajectories and avoiding obstacles. However, when reflective foil was placed on a wall in Section~1 (labeled in Figure~\ref{fig:full-track}), its predictions collapsed: steering commands became inconsistent, and the vehicle ultimately crashed into the barrier (Figure~\ref{fig:tln}). Quantitatively, \mbox{TinyLidarNet}’s steering smoothness degraded severely: the maximum difference between consecutive steering commands dropped from 0.23 (foil covered) to 0.09 (foil exposed). The low value indicates the network was repeating nearly identical predictions, effectively freezing and failing to adapt to the turn. This aligns with known limitations of LiDAR when encountering specular surfaces, which cause signal loss or spurious returns. In contrast, EARLY\textsubscript{LSTM} was unaffected by the foil and navigated the same section without difficulty. These results validate our choice of RGB-D sensing as a more robust alternative for our scenario.

\subsection{Comparison with Human Expert Driver}
\begin{figure}[bt]
\centering
\includegraphics[trim={8.75cm 4.25cm 7cm 8.75cm},clip, height=2.5cm]{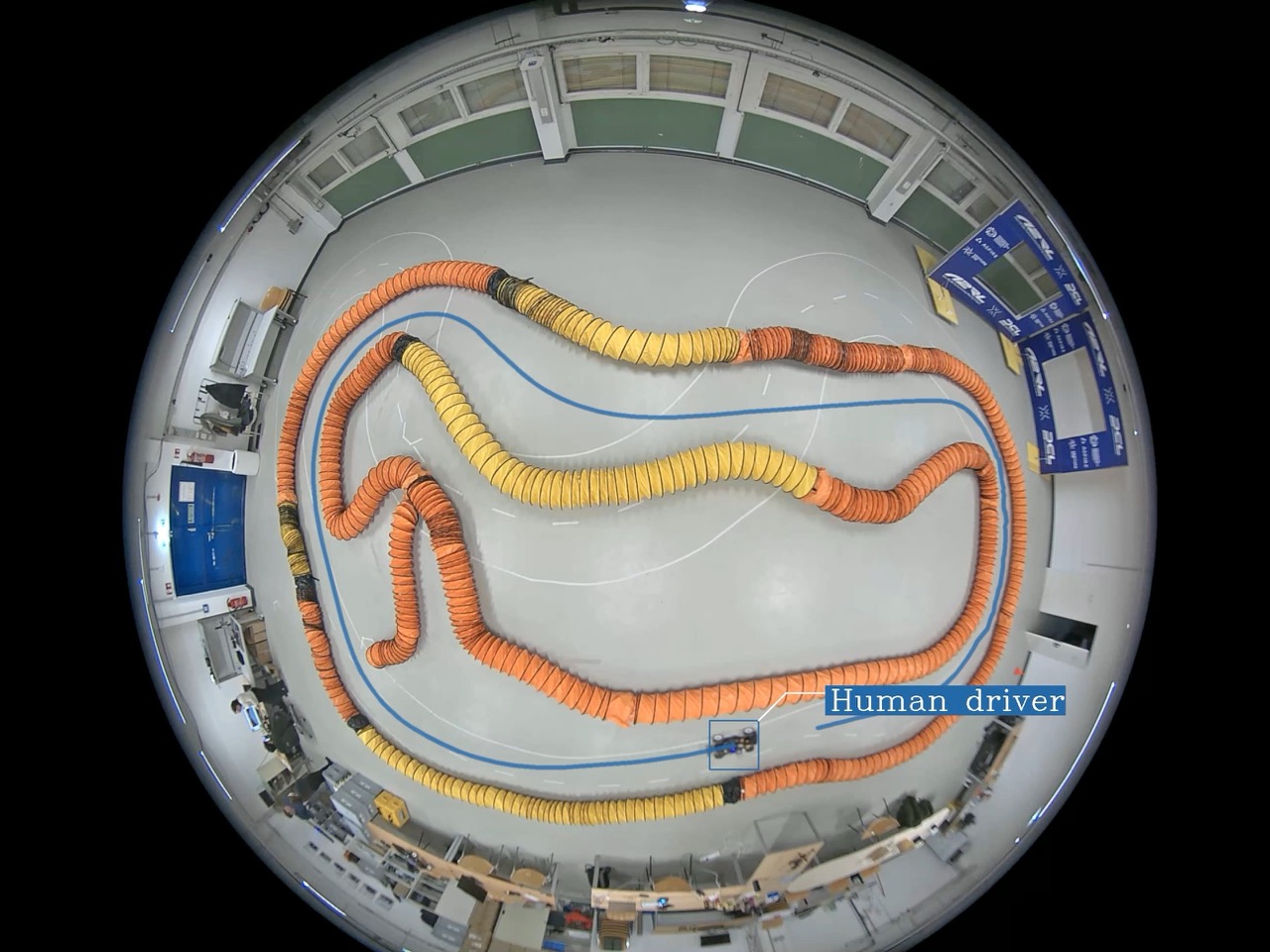}
\includegraphics[trim={7cm 3.25cm 6cm 7cm},clip, height=2.5cm]{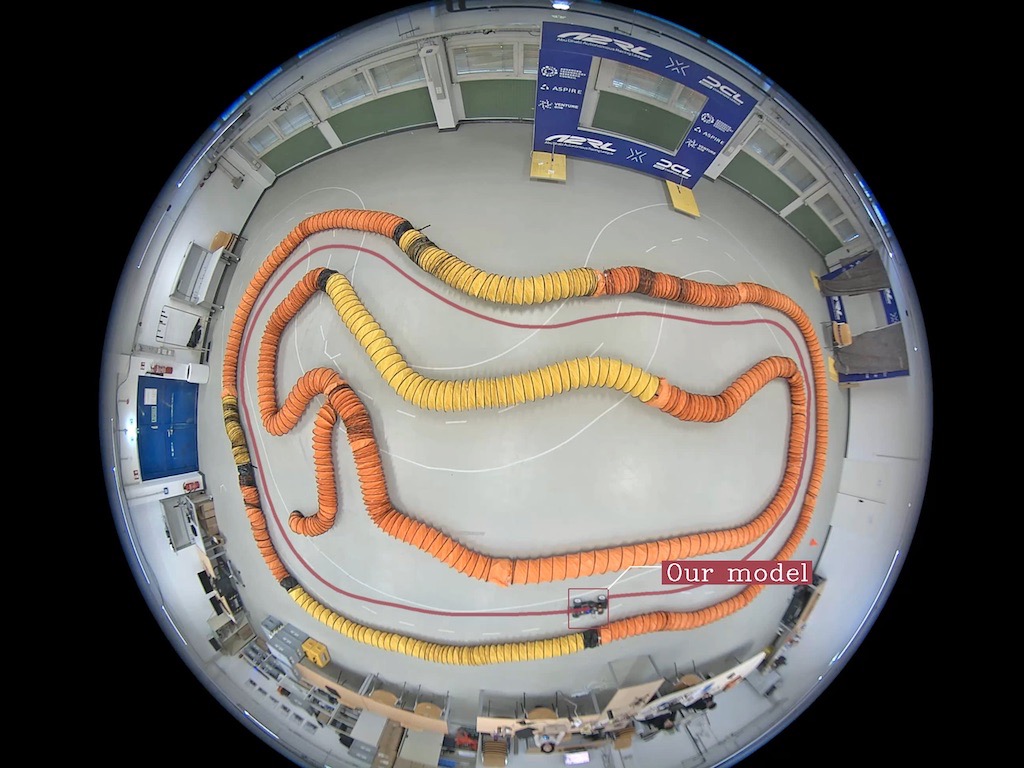}
\caption{The trajectory of the human driver on the test track is shown on the left, while the trajectory of our best model, EARLY\textsubscript{LSTM}, is shown on the right, both plotted using object tracking. While the model exhibits a smooth trajectory, it does not fully copy the driving style of the human driver.}
\label{fig:domemodel_and_human}
\vspace{-2ex}
\end{figure}

\begin{figure}[bt!]
    \centering
    \includegraphics[width=0.49\linewidth]{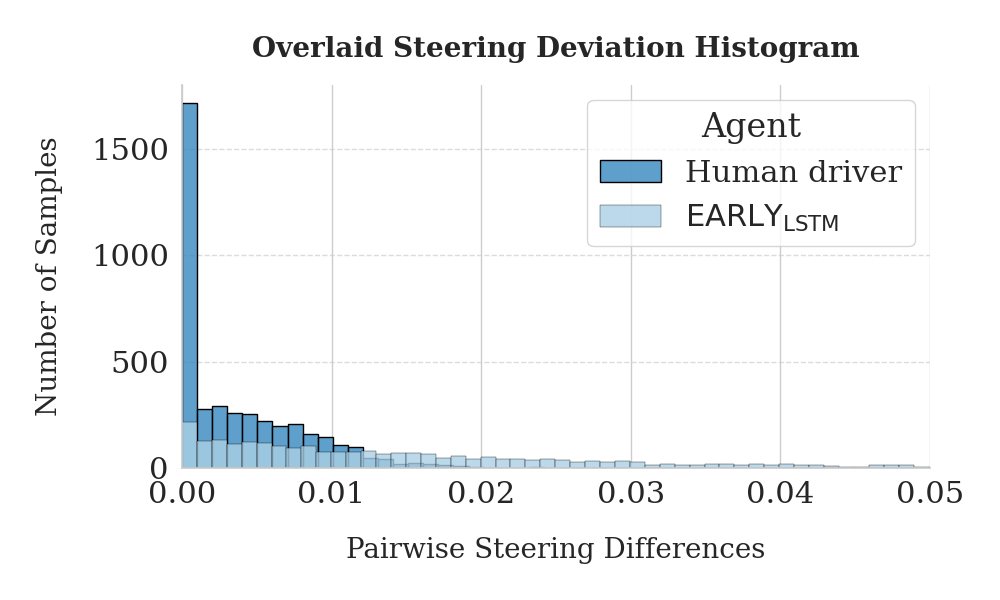}
    \includegraphics[width=0.49\linewidth]{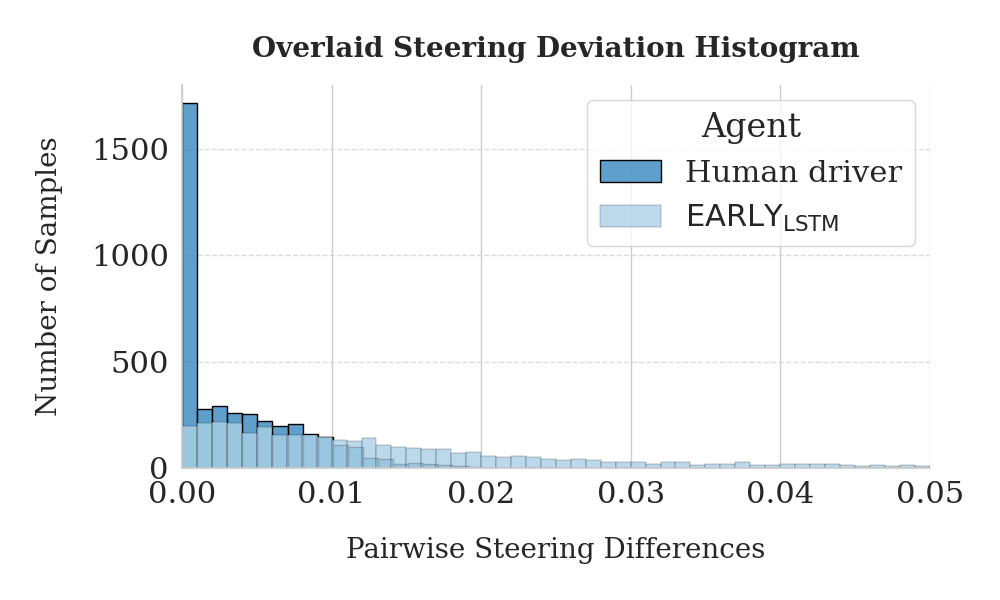}
    \caption{
    Histogram of steering differences between consecutive predictions, illustrating steering smoothness. Values near zero indicate smoother steering. Left: real-time inference (\mbox{$\approx7$fps} dropped due to hardware limits). Right: post-hoc inference. Similar distributions suggest frame drops had minimal effect on overall behavior.
    %Histogram of steering differences between consecutive predictions, showing the smoothness of steering behavior. Values closer to zero indicate smoother steering. The plots compare the distribution between model predictions and human-commanded steering during closed-loop testing. On the left, the histogram represents real-time inference where \mbox{$\approx7$fps} were being dropped due to hardware limitations. On the right, we plot the post-hoc inference results. The similarity between the two distributions suggests that frame drops did not significantly affect the model’s overall behavior.
    %, whose distribution resembles the human driver's.
    }
    \label{fig:lipschitz}
    \vspace{-3ex}
\end{figure}
After successfully deploying and testing the EARLY\textsubscript{LSTM} model under various noisy conditions, we raised the question of \textit{how this model compares to human expert driver behavior}. To explore this, we recorded a driving session of 5 continuous laps with the human driver on Map~\ref{fig:mapactive}. 

To compare the driving trajectory of the human driver and the EARLY\textsubscript{LSTM} model, we implemented an object tracking algorithm using OpenCV~\cite{opencv_library} that tracks the agent over time and plots its trajectory on the static test layout (Fig.~\ref{fig:domemodel_and_human}). Although the model produces a smooth trajectory, it does not fully replicate the human driver's style. We believe this difference is largely due to the limited amount of expert data available for training. With additional demonstrations from the same driver, the model would likely better capture the nuances of their behavior. Nevertheless, even with only a few minutes of driving data per track, the model already demonstrates effective closed-loop decision-making, highlighting its ability to generalize beyond the training distribution.

% \subsection{Impact of Frame Drops on Model Performance: Real-time vs. Post-hoc Inference Comparison}
\subsection{Impact of Frame Drops on Model Performance}
During real-time inference, our model experienced frame drops due to hardware limitations, resulting in a drop of approximately 7fps.
%one thousand missing RGB-D frames (out of a total of $4,071$, at 30fps). 
Despite this, the model successfully completed 5 fully autonomous laps on Map~\ref{fig:mapactive}, including three test scenarios with added noise, and multiple configurations and scenarios on Map~\ref{fig:full-track}. 

To assess whether these frame drops influenced the model’s behavior, we reprocessed the recorded data offline on a virtual machine, ensuring that all frames were used for inference. Figure~\ref{fig:lipschitz} compares the histogram of pairwise absolute steering differences on Map~\ref{fig:mapactive} between the model and the human driver under both conditions: (1) real-time inference with dropped frames and (2) post-hoc inference using all recorded frames. The similarity between the distributions suggests that even frame drops did not introduce significant erratic behavior or deviation in the model’s predictions.

% Additional configurations of the recurrent network could have been explored. However, our paper primarily focuses on feature extractors and only uses recurrent networks as part of the pipeline for making sequential predictions.

\section{CONCLUSION}\label{sec:conclusion}
Overcoming the sim-to-real gap for autonomous agents poses many challenges. In our paper, we provided an in-depth analysis of our successful deployment of a multimodal RGB-D recurrent controller. During deployment, we found that depth information is crucial -- the unimodal RGB feature extractor was insufficient for autonomous navigation, whereas we could successfully deploy a controller for each multimodal RGB-D feature extractor. Our early fusion LSTM agent maintained autonomy in previously unseen scenarios and high resilience to data loss and increased input noise. In order to interpret the model behavior, we employed various numerical and visual techniques, which reinforced our observations from the closed-loop tests.

% \addtolength{\textheight}{-12cm}   % This command serves to balance the column lengths
                                  % on the last page of the document manually. It shortens
                                  % the textheight of the last page by a suitable amount.
                                  % This command does not take effect until the next page
                                  % so it should come on the page before the last. Make
                                  % sure that you do not shorten the textheight too much.

%%%%%%%%%%%%%%%%%%%%%%%%%%%%%%%%%%%%%%%%%%%%%%%%%%%%%%%%%%%%%%%%%%%%%%%%%%%%%%%%

%%%%%%%%%%%%%%%%%%%%%%%%%%%%%%%%%%%%%%%%%%%%%%%%%%%%%%%%%%%%%%%%%%%%%%%%%%%%%%%%

%%%%%%%%%%%%%%%%%%%%%%%%%%%%%%%%%%%%%%%%%%%%%%%%%%%%%%%%%%%%%%%%%%%%%%%%%%%%%%%%
% \section*{APPENDIX}\label{sec:appendix}

%\section*{ACKNOWLEDGMENT} %NOT FOR THE DOUBLE BLIND SUBMISSION
%M.F. and R.G. have received funding from the European Union’s Horizon 2020 research and innovation programme under the Marie Skłodowska-Curie grant agreement No 101034277. We thank Mihai-Teodor Stănușoiu for his assistance in the data collection as our human driver.
%%%%%%%%%%%%%%%%%%%%%%%%%%%%%%%%%%%%%%%%%%%%%%%%%%%%%%%%%%%%%%%%%%%%%%%%%%%%%%%%

\bibliographystyle{IEEEtran}
\bibliography{IEEEabrv,root_short}
\end{document}